\documentclass[sigconf]{acmart}

\AtBeginDocument{%
  \providecommand\BibTeX{{%
    \normalfont B\kern-0.5em{\scshape i\kern-0.25em b}\kern-0.8em\TeX}}}

\copyrightyear{2024} 
\acmYear{2024} 
\setcopyright{rightsretained} 
\acmConference[KDD '24]{Proceedings of the 30th ACM SIGKDD Conference on Knowledge Discovery and Data Mining}{August 25--29, 2024}{Barcelona, Spain}
\acmBooktitle{Proceedings of the 30th ACM SIGKDD Conference on Knowledge Discovery and Data Mining (KDD '24), August 25--29, 2024, Barcelona, Spain}\acmDOI{10.1145/3637528.3671656}
\acmISBN{979-8-4007-0490-1/24/08}

\usepackage{CJKutf8}

\usepackage{multirow} 
\usepackage{makecell} 

\usepackage{longtable} 
\usepackage{tablefootnote}

\usepackage{pifont} 

\usepackage[shortcuts]{extdash} 
\usepackage{balance}

\begin{document}

\title{Face4RAG: Factual Consistency Evaluation for Retrieval Augmented Generation in Chinese}

\settopmatter{authorsperrow=4} 

\author{Yunqi Xu}
\affiliation{%
  \institution{Ant Group}
  \city{Shanghai}
  \country{China}
}

\author{Tianchi Cai}
\affiliation{
  \institution{Ant Group}
    \city{Hangzhou}
  \country{China}
}

\author{Jiyan Jiang}
\affiliation{
  \institution{Tsinghua University}
    \city{Beijing}
  \country{China}
}

\author{Xierui Song}
\affiliation{
  \institution{Ant Group}
    \city{Hangzhou}
  \country{China}
}

\renewcommand{\shortauthors}{Yunqi Xu, Tianchi Cai, Jiyan Jiang, and Xierui Song}

\begin{abstract}

The prevailing issue of factual inconsistency errors in conventional Retrieval Augmented Generation (RAG) motivates the study of Factual Consistency Evaluation (FCE). Despite the various FCE methods proposed earlier, these methods are evaluated on datasets generated by specific Large Language Models (LLMs). Without a comprehensive benchmark, it remains unexplored how these FCE methods perform on other LLMs with different error distributions or even unseen error types, as these methods may fail to detect the error types generated by other LLMs. To fill this gap, in this paper, we propose the first comprehensive FCE benchmark \emph{Face4RAG} for RAG independent of the underlying LLM. Our benchmark consists of a synthetic dataset built upon a carefully designed typology for factuality inconsistency error and a real-world dataset constructed from six commonly used LLMs, enabling evaluation of FCE methods on specific error types or real-world error distributions. On the proposed benchmark, we discover the failure of existing FCE methods to detect the logical fallacy, which refers to a mismatch of logic structures between the answer and the retrieved reference. To fix this issue, we further propose a new method called \emph{L-Face4RAG} with two novel designs of logic-preserving answer decomposition and fact-logic FCE. Extensive experiments show L-Face4RAG substantially outperforms previous methods for factual inconsistency detection on a wide range of tasks, notably beyond the RAG task from which it is originally motivated. Both the benchmark and our proposed method are publicly available.\footnote{\url{https://huggingface.co/datasets/yq27/Face4RAG}\label{link_face4rag}}
\end{abstract}



\maketitle

\section{Introduction}

\begin{figure*}  
\centering 
\includegraphics[width=\textwidth]{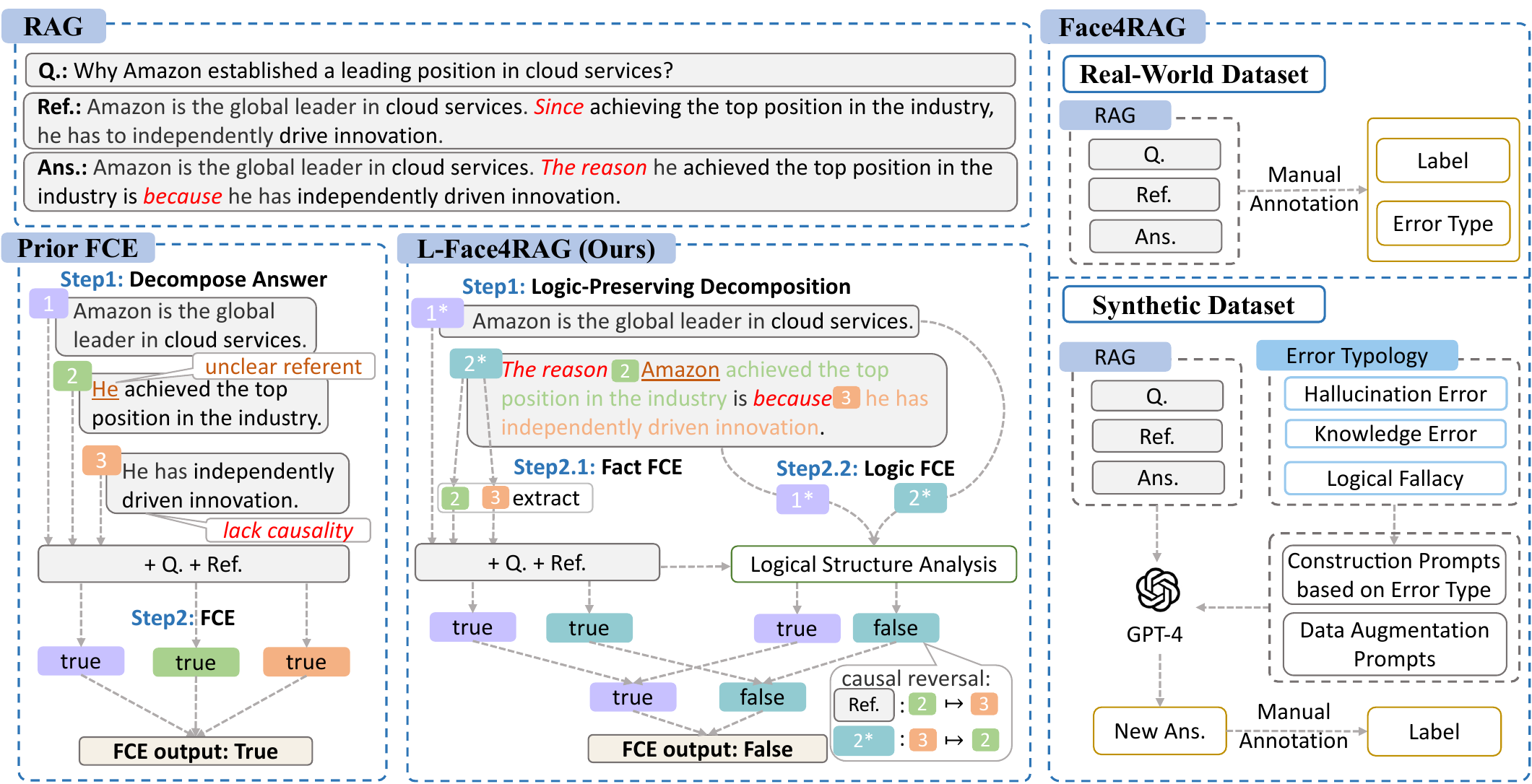}
\caption{An overview of our proposed FCE benchmark and method, in comparison with prior works. The upper left plot gives an example from RAG task. The lower left plot demonstrates previous FCE method, and the lower middle plot depicts our proposed FCE method L-Face4RAG. The upper right plot shows the procedure of constructing the real-world dataset in our proposed Face4RAG benchmark, which follows the procedure of previous benchmark. The lower right plots illustrates the construction of the synthetic dataset in the Face4RAG benchmark.}
\vspace{-2mm}
\label{Figure:1}
\end{figure*}


Retrieval Augmented Generation (RAG), a technique of augmenting the context of Large Language Models (LLMs) with relevant passages retrieved from external retrievers or search engines \cite{lewis2020retrieval}, has demonstrated strong performance on various knowledge intensive tasks such as open domain conversation \cite{thoppilan2022lamda,shuster2022blenderbot} and question answering \cite{izacard2020leveraging}. Despite its bright prospect, factual consistency remains a critical issue for RAG systems. Recent assessment reveals that even for the leading-edge commercial RAG systems like Bing Chat and Perplexity, barely over half of their outputs are factual consistent with the references \cite{liu2023evaluating}. This issue urges the need of studying factual consistency evaluation (FCE) in the RAG task.

Various FCE methods have been proposed to evaluate the factual consistency of specific RAG systems, among which a two-step approach shows promising results, especially for evaluating long answers \cite{min2023factscore,chern2023factool}. As shown in the bottom left of Figure \ref{Figure:1}, this approach first segments the answer into shorter pieces, then evaluates the factual consistency of each segment with respect to the given reference. In this way, the evaluation of a long answer is decomposed into evaluations on several simpler pieces of information, which improves the detection of factual inconsistency. 


In previous works, these FCE methods are evaluated by answers generated by the underlying LLM in the specific RAG system being studied \cite{es2023ragas,hu-etal-2023-refchecker}. Despite their effectiveness on the specific system, it is unclear how these methods generalize to new RAG systems. As discovered in a recent study \cite{min2023factscore}, the optimal FCE method may vary when evaluating different LLMs, hence achieving a superior performance regarding some certain LLM does not guarantee a strong performance on other LLMs. In this sense, previous benchmarks generated by a single LLM are not fair enough to evaluate the overall performance of FCE methods.


To fill this gap, in this paper, we first construct a comprehensive benchmark to enable the evaluation of FCE methods independent of the underlying LLM. Specifically, we first propose a novel error typology to cover various factual consistency errors in RAG, which includes three main categories, i.e., hallucination error, knowledge error, and logical fallacy, and is further divided into nine error types. Based on our predefined error typology, we construct a synthetic dataset in Chinese to assess the effectiveness of FCE methods across the different types of errors. Furthermore, we construct a real-world dataset in Chinese by generating answers using six distinct LLMs within RAG tasks. Empirical analysis on the real-world dataset shows that 6.96\% of all factual inconsistent samples involve logical fallacies. In addition, we observe that different LLMs exhibit diverse error distributions, which echoes previous research \cite{min2023factscore} and justifies our motivation of constructing a comprehensive benchmark independent of LLMs.

While logical fallacy accounts for a considerable proportion of factual inconsistency errors in the real-world dataset, existing FCE methods may be incapable of detecting these sophisticated errors involving logical connections among multiple text segments, since the decomposition step may neglect the logical connections between segments in the original answer. Figure \ref{Figure:1} provides a showcase where a careless decomposition may mistakenly remove the cause-effect relation, leading to a wrong evaluation result. 

To resolve this issue, we develop the Logic-Enhanced FActual Consistency Evaluation for RAG (L-Face4RAG) method to better handle the logical consistency in the RAG task. L-Face4RAG has two core designs, i.e., logic-preserving answer decomposition and fact-logic FCE. Specifically, in the answer decomposition step, we propose three principles for decomposition based on semantic linkages and logical connections. We design an elaborated prompt accordingly and construct few-shot examples to help LLM better follow the above principles. In the subsequent FCE step, we assess the factual consistency of each segment from two perspectives, i.e., the \textit{fact consistency} and \textit{logical consistency}. The former perspective aims to detect hallucination or knowledge errors, while the later is responsible for the logical fallacy errors. We further design a chain-of-thought (COT) \cite{wei2022chain} prompt at each stage to instruct the LLM to better handle the inconsistency error in a step-by-step manner. Figure~\ref{Figure:1} gives a detailed demonstration of L-Face4RAG. 

Finally, we conduct extensive experiments to verify the effectiveness of L-Face4RAG. Compared to previous FCE methods for RAG, L-Face4RAG attains substantially higher accuracy on both synthetic and real-world datasets, regardless of the error type or underlying LLM. Notably, although it is motivated by FCE in Chinese RAG, its superiority is consistent on other FCE tasks. Specifically, additional experiments on English FCE benchmarks for RAG\cite{es2023ragas,hu-etal-2023-refchecker}, summarization\cite{pagnoni2021understanding, fabbri2021summeval}, dialogue\cite{honovich2021q,gupta2021dialfact} and fact verification\cite{schuster2021get} show that L-Face4RAG achieves SOTA on most of the tasks (6 out of 7), as well as a substantially higher averaged score. We further conduct ablation studies to verify the core designs of L-Face4RAG, i.e., the logic-preserving answer decomposition approach and two-stage consistency evaluation with carefully designed COT prompts. 

The contributions of this work are summarized as follows:
\begin{itemize}
\item We construct the first comprehensive FCE benchmark in RAG, the Face4RAG, which includes a carefully designed error typology, a synthetic dataset, and a real-world dataset. Face4RAG allows to evaluate FCE method on specific error types or various real-world error distributions. 
\item We propose a new FCE method called L-Face4RAG with two novel designs of logic-preserving answer decomposition and fact-logic FCE to better detect the logic fallacy in the examined answer. 
\item Extensive experiments justify the proposed error typology, evaluate the effectiveness of L-Face4RAG on a wide range of FCE tasks, and provide further insights on the distinct error type distributions of various LLMs. All datasets and method are released for better reproducibility. 
\end{itemize}

%
%
%
\section{Related Work}
\textbf{Traditional FCE Methods}.
Evaluating the factuality of model generated results is widely studied across various language model generation domains like text summarization \cite{gao2023human}, dialogue summary \cite{zhu2023annotating} and question-answering \cite{bai2023benchmarking}. When the golden labels are given, prior methods using exact match metrics \cite{izacard2020leveraging,lewis2020retrieval,chen2017reading} or similarity-based metrics are proposed \cite{zhang2019bertscore,chen2019evaluating}. However, high quality answers can vary a lot, hence these approaches using golden labels may significantly underestimate the models' performances, especially for long answers \cite{wang2023evaluating}. 

\textbf{FCE for Long-form Answers.} To effectively evaluate factuality of long answers, recent FCE research mostly take a two step approaches \cite{kamoi2023wice,lattimer2023fast}, where in the first step, the long-form answer is decomposed into shorter segments, such as sentences \cite{kryscinski2019evaluating,lattimer2023fast}, sub-claims \cite{kamoi2023wice,chern2023factool}, individual facts \cite{min2023factscore} and structured triplets \cite{hu-etal-2023-refchecker}. Then the second step evaluates the verifiability of each segment with respect to the given reference text \cite{laban2022summac,lattimer2023fast,zha2023alignscore}, which can be efficiently done by modern general purpose LLM  \cite{chen2023felm,min2023factscore}, e.g., GPT-4. Although we follow the two-step approach, our method differs from them in the ability of leveraging logical connections via special designs of logic-preserving decomposition and fact-logic FCE. 

\textbf{FCE Benchmarks.} Prior benchmarks for FCE mostly focus on specialized tasks like summarization \cite{kryscinski2019evaluating,tang2022understanding,es2023ragas}. For FCE in RAG, existing benchmarks are derived from specific LLMs, such as Refchecker \cite{hu-etal-2023-refchecker} and FELM \cite{chen2023felm}, which are constrained by the error type distribution of the underlying LLMs. Unlike these benchmarks, we construct a synthetic dataset based on our error typology, which enables evaluation independent to any underlying LLM.

\section{FACE4RAG Benchmark}



\begin{figure}[tb]
  \centering
  \includegraphics[width=\linewidth]{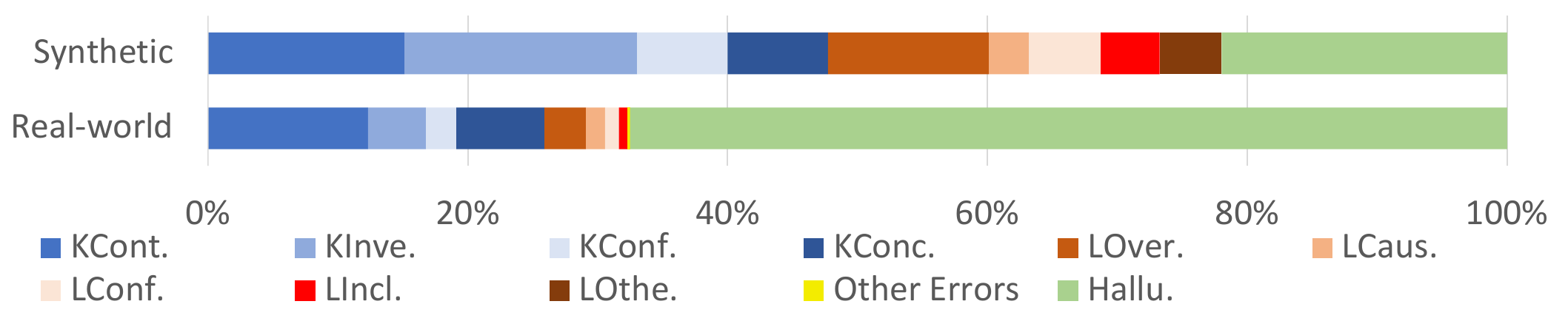}
  \vspace{-6.5mm}
  \caption{
  Error type distribution of factually inconsistent samples in the two datasets of our Face4RAG benchmark.}
  \label{fig:error typology distribution of benchmark}
  \vspace{-5mm}
\end{figure}

Recall that existing FCE benchmarks only use answers generated by some certain LLMs, which may fail to evaluate FCE methods on other LLMs with different error distributions or unseen error types. 
To remedy this issue, in this section, we propose a novel approach to construct a FCE benchmark for RAG, which is independent of the underlying LLMs and called \textit{FActual Consistency Evaluation for RAG} (Face4RAG). Face4RAG contains an error-type-oriented synthetic dataset and a real-world dataset. To construct the synthetic dataset, inspired by the error typology used in an exam designed for humans, i.e. the National College Entrance Examination of China, we first propose a novel error typology to classify any factual consistency error in RAG task, which includes nine types of errors belonging to three main categories. Based on the proposed error typology, we then construct a synthetic dataset to evaluate FCE methods on each type of the error. Besides the synthetic dataset, we also collect samples from six commonly used LLMs to construct a real-world benchmark, which aims to evaluate the overall factual consistency of FCE methods in real-world scenarios
. The details about Face4RAG can be seen in Table~\ref{tab:dataset Statistics} and Figure~\ref{fig:error typology distribution of benchmark}.

\begin{table}[tb]
\caption{ Statistics of the synthetic and real-world datasets in the Face4RAG benchmark. For each dataset, the answer-level and segment-level statistics on the number of samples, the average sample length in terms of characters and the rate of positive samples are reported.}
\vspace{-1mm}
\label{tab:dataset Statistics}
\begin{tabular}{cccccc}
\toprule
\multirow{2}{*}{\textbf{Statistics}} & \multicolumn{2}{c}{\textbf{Synthetic Dataset}} && \multicolumn{2}{c}{\textbf{Real-world Dataset}} \\ \cline{2-3}\cline{5-6}
& Answer & Segment && Answer & Segment \\ \midrule
Num. Samples & 1299 & 6737 && 1200 & 6143 \\
Avg. Length & 289.3 & 45.4 && 307.7 & 45.2 \\
Positive Rate & 30.3\% & 55.8\% && 63.3\% & 85.6\% \\
\bottomrule
\end{tabular}
\vspace{-3mm}
\end{table}

\subsection{Error Typology in FCE} \label{error_typology}



Our error typology for FCE is inspired by the questions in the National College Entrance Examination of China \cite{gaokao2019chinese}, which are carefully designed to test the ability of human to evaluate factual consistency. In this examination, reading comprehension is a major section to evaluate the participants' skill of understanding a Chinese text. Factual consistency evaluation is a typical task in this section. Given the text, the participants are required to evaluate the correctness of several answers to a specific question \cite{gaokao2019chinese}, which is essentially a RAG task (see examples in Table~\ref{Example of the Reading Comprehension Section} in the appendix).
As these questions are designed for a competitive entrance examination of higher education institutions at the undergraduate level, they are generally hard to answer and cover a wide range of factual inconsistency error types. Accordingly, we develop a novel error typology for RAG, which comprises three main categories and is further classified into nine error types. In the following, we give a detailed description of our proposed error typology.

\textbf{Hallucination Error} This class of error refers to the situation when the answer contains information that cannot be traced back to the reference \cite{maynez2020faithfulness}. Note that there are two main usages of the term hallucination in previous literature: one refers to "unfaithful or nonsensical" generated answers \cite{ji2023survey}, the other further includes "unverifiable" answers using the given context \cite{maynez2020faithfulness,thomson2020gold}. Here we adopt the second usage that has a larger scope. 
\begin{itemize}
    \item \textit{Hallucination Error (Hallu.)} refers to the situation when the answer is either unfaithful or unverifiable using the given context (even when it is factually correct).
\end{itemize}
%
%
%

\textbf{Knowledge Error} This class of error refers to the situation when the information contained by the answer is inaccurate or incorrect regarding the reference \cite{chen2023felm}. This may occur in various components of a sentence, such as the subjects, predicates, objects, adverbials of time and place, etc. We classify the knowledge error into four error types:
\begin{itemize}
\item\textit{Contradiction Error (KCont.)} refers to the situation when the statement in the answer conflicts with information from the reference.

\item\textit{Entity Inversion Error (KInve.)} refers to the situation when entities in the answer (events, processes, or concepts) are swapped in their positions as compared to the reference.

\item\textit{Conflation Error (KConf.)} refers to the situation when the entities in the reference (subjects, predicates, or objects) are inaccurately combined, altering the original meaning.

\item\textit{Conceptual Substitution Error (KConc.)} 
refers to the situation when a term or concept in the reference is erroneously replaced by a different (though possibly related) concept.
\end{itemize}

\textbf{Logical Fallacy} This kind of error occurs when the answer contains statements that are either without logical support or have a logical relation that conflicts with the information from the reference. This incongruity undermines the logical validity of the answer with respect to the reference, which results in an unsound argument or misleading information \cite{jin2022logical,petric2020logical}.  We further divide the logical fallacy into four error types:
\begin{itemize}
\item\textit{Overgeneralization Error (LOver.)} refers to the situation when a specific detail or attribute from the reference is incorrectly applied to a broader category or group in the answer. 

\item\textit{Causal Confusion Error (LCaus.)} refer to reverse the cause and effect in the reference, or mistakenly adding a causal relationship between two noncausal segments.

\item\textit{Confusing Sufficient and Necessary Conditions Error (LConf.)} is the case when the necessary conditions in reference are misinterpreted as sufficient and necessary conditions.
\item\textit{Inclusion Relation Error (LIncl.)} is the case where statements that are unrelated or have certain relationship except inclusion in the reference are misrepresented in the answer to have an inclusion relationship (e.g., hierarchical or subset).
\end{itemize}
\vspace{-1mm}

For each of the error types defined above, we provide several examples to help better understand and distinguish it from other types. See detailed examples in Table ~\ref{tab:negative}.

\begin{CJK*}{UTF8}{gbsn}
\begin{table*}
    \caption{Examples for Knowledge Error and Logic Fallacy. For each error types,  the example in Chinese and the translated version to English are presented. The colored text spans highlight the segments of factual inconsistency errors.} 
    \vspace{-1mm}
    \label{tab:negative}
    \begin{tabular}{p{0.08\textwidth}p{0.42\textwidth}p{0.42\textwidth}}
    \toprule
     Error Type & Original Text & Factual Inconsistent Text\\
\midrule
KCont. &  功能饮料中的维生素、矿物质等，对于运动后快速补充身体营养，\textcolor{blue}{消除}疲劳具有一定作用。\newline The vitamins and minerals in energy drinks play a certain role in quickly replenishing nutrients and \textcolor{blue}{eliminating} fatigue after exercise. & 功能饮料中的元素、微生物等，对于运动后快速补充身体营养，\textcolor{blue}{增加}疲劳具有一定作用。\newline The vitamins and minerals in energy drinks play a certain role in quickly replenishing nutrients and \textcolor{blue}{inducing} fatigue after exercise.\\ \hline 
KInve. & 一般蚕可以活一个多月，其中从孵化到结茧根据季节不同大约是\textcolor{blue}{25-32}天，变成蛹后有\textcolor{brown}{15-18}天，最后成蛾是1-3天。\newline A typical silkworm can live for just over a month, during which the period from hatching to cocooning varies roughly from \textcolor{blue}{25 to 32} days depending on the season, followed by \textcolor{brown}{15 to 18} days as a pupa, and finally 1 to 3 days as a moth. & 一般蚕可以活一个多月，其中从孵化到结茧根据季节不同大约是\textcolor{brown}{15-18}天，变成蛹后有\textcolor{blue}{25-32}天，最后成蛾是1-3天。\newline A typical silkworm can live for just over a month, during which the period from hatching to cocooning varies roughly from \textcolor{brown}{15 to 18} days depending on the season, followed by \textcolor{blue}{25 to 32} days as a pupa, and finally 1 to 3 days as a moth. \\ \hline
KConf. & 防晒霜中的\textcolor{blue}{无机化学物质}可以\textcolor{blue}{反射或散射皮肤上的光线}，而\textcolor{brown}{有机(碳基)化学物质}可以\textcolor{brown}{吸收紫外线}。\newline The \textcolor{blue}{inorganic chemicals} in sunscreen can \textcolor{blue}{reflect or scatter light on the skin}, while \textcolor{brown}{organic (carbon-based) chemicals} can \textcolor{brown}{absorb ultraviolet rays}. & 防晒霜中的\textcolor{blue}{无机化学物质}和\textcolor{brown}{有机(碳基)化学物质}都可以\textcolor{blue}{反射或散射皮肤上的光线}、\textcolor{brown}{吸收紫外线}。\newline Both the \textcolor{blue}{inorganic chemicals} and \textcolor{brown}{organic (carbon-based) chemicals} in sunscreen can \textcolor{blue}{reflect or scatter light on the skin} and \textcolor{brown}{absorb ultraviolet rays}. \\ \hline
KConc. & 随着健康意识的增强，越来越多的人开始注重\textcolor{blue}{膳食平衡}。\newline With the increasing awareness of health, more and more people are beginning to focus on \textcolor{blue}{a balanced diet}. & 随着健康意识的增强，越来越多的人开始注重\textcolor{blue}{膳食的有机质量}。\newline With the increasing awareness of health, more and more people are beginning to focus on \textcolor{blue}{the organic quality of their diets}. \\ \hline
LOver. & 一般的我们平时见到的\textcolor{blue}{蜘蛛}都是晚上出来。\newline The \textcolor{blue}{spiders} that we usually see tend to come out at night. & 一般的我们平时见到的\textcolor{blue}{昆虫}都是晚上出来。\newline The \textcolor{blue}{insects} that we usually see tend to come out at night.\\ \hline
LCaus. & \textcolor{blue}{随着}信息技术的快速发展，大数据在各行各业中的应用越来越广泛。\newline  \textcolor{blue}{With} the rapid development of information technology, the application of big data across various industries is becoming increasingly widespread.& 大数据在各行各业中的应用越来越广泛，这\textcolor{blue}{导致}了信息技术的快速发展。\newline The application of big data across various industries is becoming increasingly widespread, \textcolor{blue}{leading to} the rapid development of information technology.\\\hline
LConf. & 为了获得某项荣誉学生奖学金，学生\textcolor{red}{必须}具备以下条件：\textcolor{blue}{成绩优秀}、\textcolor{blue}{品行端正}、\textcolor{brown}{参加社会实践活动}。\newline To receive a certain honor student scholarship, students \textcolor{red}{must} meet the following criteria: \textcolor{blue}{excellent academic performance}, \textcolor{blue}{good moral character}, and \textcolor{brown}{participation in social practice activities}.& 学生\textcolor{blue}{成绩优秀}、\textcolor{blue}{品行端正}\textcolor{red}{就可以}获得某项荣誉学生奖学金。\newline Students with \textcolor{blue}{excellent academic performance} and \textcolor{blue}{good moral character} \textcolor{red}{can} receive a certain honorary student scholarship.\\ \hline
LIncl. & 坚持锻炼身体可以提高心肺能力，加强肌肉的耐力，提高身体的抗疲劳能力。\newline Regular exercise can enhance cardiorespiratory fitness, strengthen muscle endurance, and improve the body's resistance to fatigue.& 坚持锻炼身体可以提高心肺能力，\textcolor{blue}{例如}加强肌肉的耐力、提高身体的抗疲劳能力。\newline Regular exercise can enhance cardiorespiratory fitness, \textcolor{blue}{such as} strengthening muscle endurance and improving the body's resistance to fatigue. \\ \bottomrule
\end{tabular}
\end{table*}

\subsection{Synthetic Dataset} \label{synthetic_dataset}
Based on the above proposed error typology, we construct a synthetic dataset. In the dataset, the positive samples are factual consistent, whereas each negative sample has at least one factual inconsistency error. The dataset is constructed based on WebCPM \cite{qin2023webcpm}, a web-enhanced question answering dataset in Chinese. Due to the space limit, in the following we briefly describe the process of dataset generation. Please refer to Appendix~\ref{construction of the synthetic dataset} for more details of the construction of our synthetic dataset. 

\textbf{Negative Samples} For each specific error type in the typology, we design a prompt to generate samples with this error. For the hallucination error, we setup three levels of difficulty for the evaluator to detect inconsistency and construct samples accordingly. For the remaining two categories, i.e., knowledge error and logical fallacy, we design a specific prompt for each error type except the Contradiction Error (\textit{KCont.}). Since \textit{KCont.} may occur at different levels of granularity \cite{de2008finding}, i.e., word or sentence, we design one prompt for each. Apart from the above error types, we construct a new error type called Other Logical Fallacy (LOthe.), which accounts for potential errors in some complex logical connections uncovered by our previously defined four types of logical fallacy.

\textbf{Positive Samples} To enrich the sample diversity, we apply the augmentation technique in \cite{li2022data}. Specifically, the original positive samples in WebCPM are augmented by synonym replacing and paraphrasing via certain prompt at either word or sentence level. 

\textbf{Human Annotation Refinement} To enhance the quality of the coarse labels derived above, we further engage 12 human experts to annotate the factual consistency of each answer via a two-step approach \cite{min2023factscore}.

\begin{figure}[tb]
  \centering
  \includegraphics[width=\linewidth]{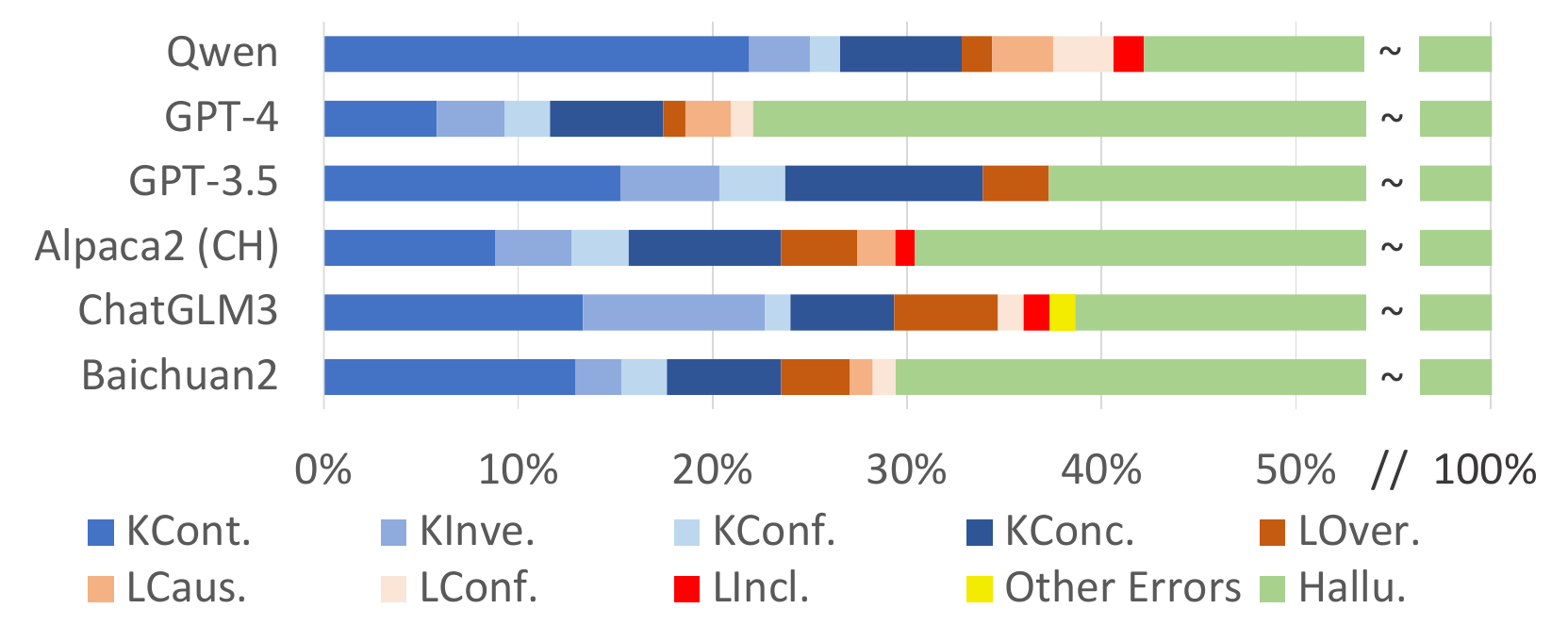}
  \vspace{-6mm}
  \caption{Error type distributions of the six LLMs in our real-world dataset (we omit the 50\%$\sim$100\% region in type ratio).}
  \label{fig:error typology distribution}
\vspace{-4mm}
\end{figure}

\subsection{Real-World Dataset}

The synthetic dataset is generated based on various predefined error types without considering the distribution of these error types in the real world.
Consequently, there is a need for another evaluation dataset that better aligns with the actual distribution of answers in real-world RAG scenarios, thus serving as a real-world dataset. In contrast to previous studies that relied solely on GPT-based LLMs for generating responses to create their evaluation sets \cite{chen2023felm,min2023factscore}, we adopt a more comprehensive approach by utilizing six different LLMs to construct our real-world dataset.

Specifically, we first collect 200 questions along with corresponding references. We then prompt six commonly used LLMs to generate answers for the questions based on the references, including gpt\-/4\-/turbo (GPT\-/4) \cite{achiam2023gpt}, gpt\-/3.5\-/turbo (GPT\-/3.5) \cite{chatgpt}, Baichuan2\-/13B\-/Chat (Baichuan2) \cite{baichuan2023baichuan2}, ChatGLM3 \cite{zeng2022glm}, Qwen\-/14B\-/Chat (Qwen) \cite{qwen}, and Chinese\-/Alpaca\-/2\-/13B\-/16k \cite{Chinese-LLaMA-Alpaca} {(Alpaca2 (CH))} . In this way, we derive a total of 1200 data points, which constitute the real-world dataset. 

For each data point, we follow the same human annotation procedure as in our synthetic dataset to inspect if it is factually consistent. Moreover, if an answer is deemed factually inconsistent, the annotator will assign a specific error type from our proposed error typology to the answer. When the annotators notice that error of the answer does not fall into the aforementioned error types, they will mark the answer as "Other Errors".


We now conduct empirical analyses on the error typology and the behaviors of various LLMs on the above real-world dataset. 

\textbf{Overall Error Type Distribution}
We first justify our study on logical fallacy consistency detection by empirically showing that the logical fallacy errors are prevailing in the answers generated by various LLMs. To this end, we analyze the distribution of the error types annotated across the entire real-world dataset. As shown in Figure~\ref{fig:error typology distribution of benchmark}, the hallucination error, knowledge error and logical fallacy account for 73.78\% , 28.31\%, 6.96\% of all the inconsistent samples, respectively. It worth note that this 6.96\% logical fallacy errors are not studied in the previous FCE methods. Besides, only 0.23\% of the inconsistent samples are marked as "Other Errors" by annotators, which suggests the comprehensiveness and completeness of our proposed error typology. 

\textbf{Error Distribution of Various Models } We then look deeper into the error types and their distributions among various LLMs in RAG. As presented in Figure~\ref{fig:error typology distribution}, various LLMs exhibit distinct distributions on error types. For instance, \textit{LIncl.} emerges in three of the LLMs, and \textit{LCaus.} and \textit{LConf.} occurs in four models. In addition, while \textit{Hallu.} exists in all models, GPT-4 has a notably high percentage, with 77.91\% of its errors being of this specific type; in comparison, Qwen only has 57.81\% Hallucination Error and a higher proportion of logical fallacy at 9.38\%. The distinct error types distributions of different LLMs suggest that FCE methods evaluated on a specific LLM may not generalize well to other LLMs, indicating the necessity for constructing a benchmark that is independent of the underlying LLM.




\begin{figure*}  
\centering 
\includegraphics[width=0.9\textwidth]{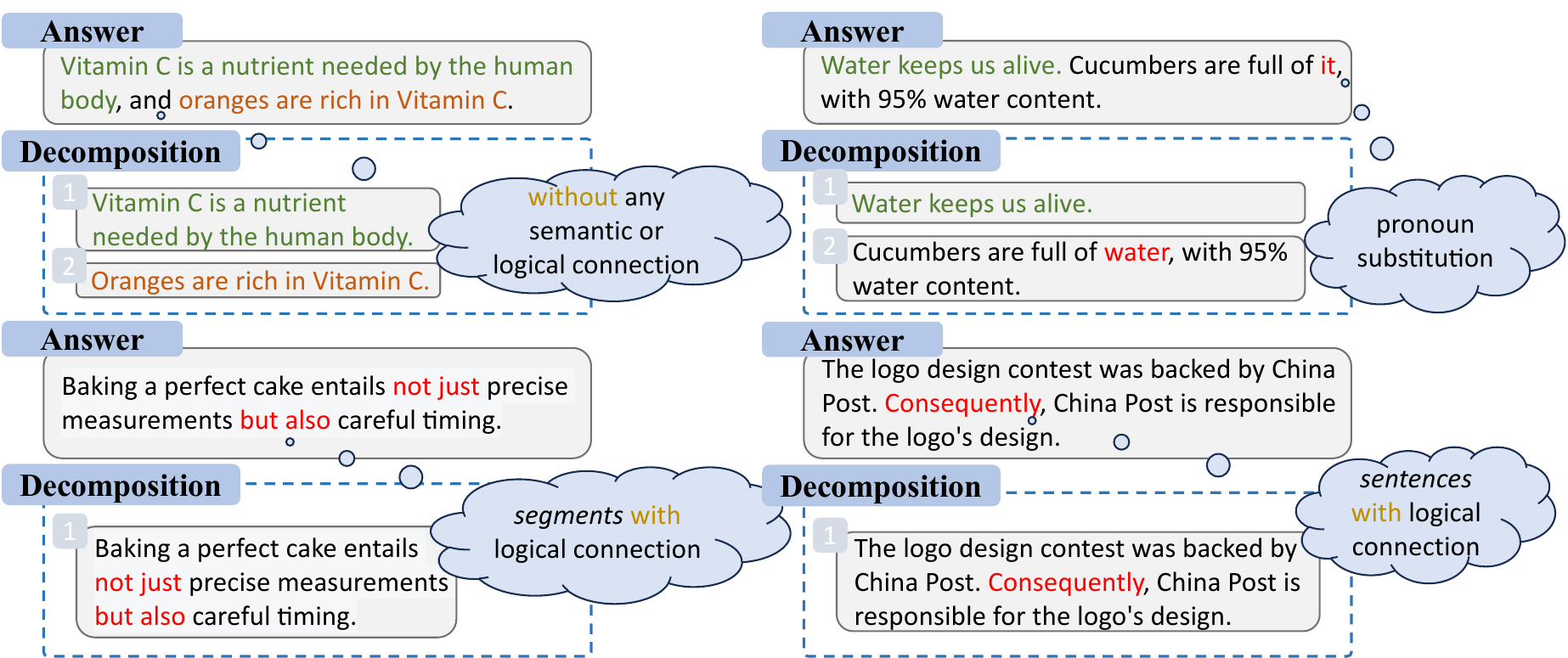}
\caption{A few examples for our proposed logic-preserving answer decomposition.}
\label{answer_decomposition}
\end{figure*}

\section{Logic-enhanced Factual Consistency Evaluation}


In the above statistic analysis, logical fallacy accounts for a considerable proportion of factual errors in real-world RAG scenarios. However, as we have analyzed before, existing FCE pipelines neglect the logical connections between segments in the original answer, which may result in wrong factual consistency evaluation result for samples with logical fallacy. Hence, to improve the evaluation ability of factuality consistency, a natural direction is to design an advanced FCE method that is capable of handling logical connections in long answers. 

In this section, we propose a novel pipeline called {\textit{Logic-Enhanced FActual Consistency Evaluation for RAG (L-Face4RAG)}}, which explicitly takes logical connections into consideration when evaluating the factual consistency. L-Face4RAG has two core modules, i.e., logic-preserving answer decomposition and fact-logic FCE, which will be described as follows.

\subsection{Logic-Preserving Answer Decomposition}
Most existing studies directly decompose answers into segments, each containing only a single piece of information \cite{min2023factscore,kamoi2023wice}. In contrast, we propose to decompose the answers based on semantic linkages\footnote{Semantic linkage refers to the connection or association between different pieces of information based on their meanings or semantic content \cite{jackendoff1992semantic}.} and logical connections, which preserves the logical relationships and facilitates logical consistency evaluation. The core design in this module is an elaborated prompt based on the following three principles for answer decomposition. 

\begin{itemize}
    \item We prompt GPT-4 to execute the decomposition only when the two or multiple sentences do not exhibit strong semantic or logical connection.
    \item To ensure that each segment can be understood by GPT-4 independently without leveraging other segments, any pronoun in a segment that refers to other contextual information should be substituted with appropriate reference. 
    \item During the decomposition process, GPT-4 is required to maintain the sentence structure of the original answer to the best extent. This principle alleviates the risk of introducing additional hallucination to the original answer.
\end{itemize}

In order to help GPT-4 better understand our principles for answer decomposition and deal with texts with various formats, we construct three kinds of instances to serve as the few-shot examples. The specific type of instances are as follows:
\begin{itemize}
    \item \textit{Logical Connection} refers to the instances having logical connections between the sentences and thus, GPT-4 needs to learn the solution of the logical connections during the decomposition process.
    \item \textit{Pronoun Substitution} involves replacing pronoun with their referent entities during the answer decomposition to make each answer segment understandable on its own, without reliance on other segments.
    \item \textit{Unique Format} refers to the instances with unique format and may be difficult for GPT-4 to decompose properly.
\end{itemize}


Examples of the answer decomposition are provided in Figure~\ref{answer_decomposition} and the detailed prompt is provided on our benchmark webpage.\textsuperscript{\ref{link_face4rag}}

\subsection{Fact-Logic FCE}
Previous methods directly invoke an LLM to evaluate the decomposed segments and overlook the logical fallacy. To evaluate the logical fallacy, we develop a two-stage procedure for factual consistency evaluation, which consists of a conventional stage of fact consistency evaluation and an extra stage that evaluates from both perspectives of fact and logic; we introduce the COT mechanism \cite{wei2022chain} into both stages to improve LLM's ability of evaluation. The prompts for each stage are provided at our benchmark webpage.\textsuperscript{\ref{link_face4rag}}

\textbf{Fact Consistency Evaluation} In this stage, GPT-4 is instructed to assess the consistency of each piece of information in the segment against the reference, which mainly concerns with the hallucination error and the knowledge error. Unlike previous methods that directly instruct the model to assess the consistency with the reference \cite{es2023ragas,chen2023felm}, we use the COT technique to guide the model to evaluate the segment step-by-step, with the following steps:

\begin{enumerate}
    \item \textit{Informational Points Extraction}: GPT-4 extracts all informational points from the segment. This step ensures that each component of the segment will be evaluated.
    \item \textit{Context Identification}: For each informational point, GPT-4 locates the corresponding content within the reference.
    \item \textit{Fact Consistency Check}: GPT-4 conducts a thorough fact consistency check for each informational point against its corresponding context. A segment is deemed consistent if and only if every single informational point aligns fact consistent with the reference. 
\end{enumerate}


The above instruction imposes GPT-4 to evaluate consistency with each relevant content rather than the full context of the reference, reducing the probability of misjudging positive samples. We will empirically justify this point in our experiments.

\begin{figure}
  \centering
  \includegraphics[width=\linewidth]{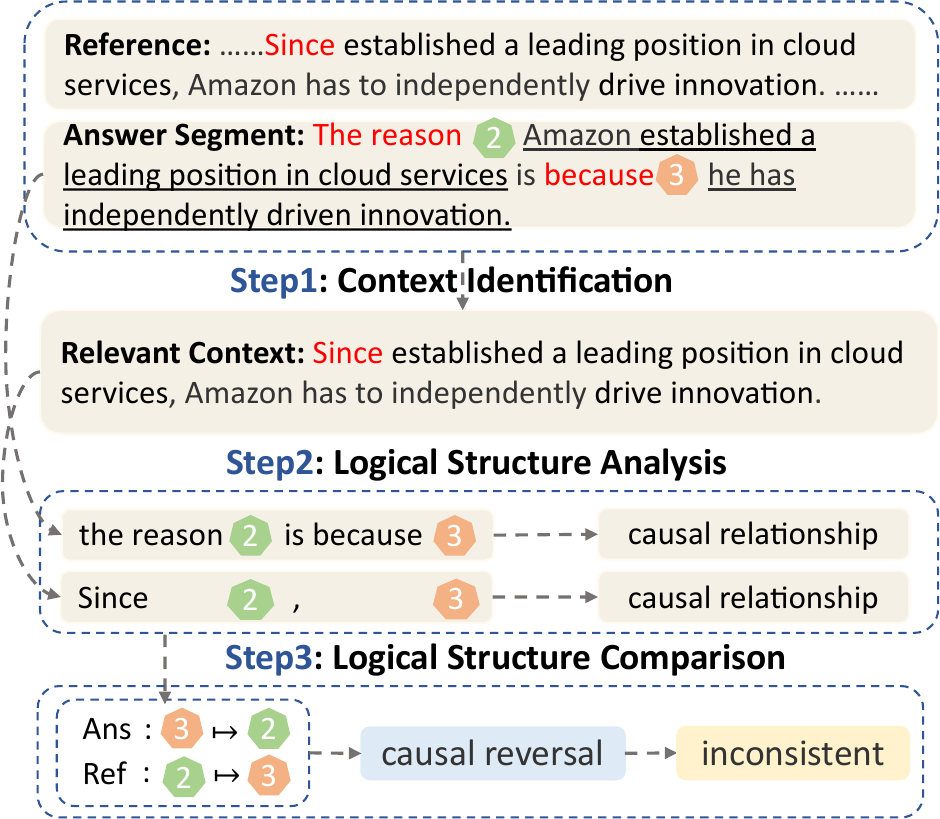}
  \caption{The process of logic consistency evaluation.}
  \label{fig:logic fce}
\end{figure}

\textbf{Logic Consistency Evaluation} In this stage, GPT-4 is instructed with a COT prompt to evaluate the logical fallacy. Since no FCE method has explicitly handled logical fallacy before, this is a novel stage for FCE, in which we elaborate a COT prompt as follows. {The specific process is shown in Figure~\ref{fig:logic fce}.}
\begin{enumerate}
    \item \textit{Context Identification}: Given an answer segment, GPT-4 identifies its relevant context from the reference.
    \item \textit{Logical Structure Analysis}: GPT-4 then analyzes the logical structure for both answer and relevant context.
    \begin{enumerate}
        \item Identify the logical connections, and the sentence components connected by these logical connections. 
        \item Determine the type and function of logical connections to understand the logical structure between sentence components, e.g., causal, conditional, etc. 
        \item Map sentence components to their corresponding logical relations, e.g., cause and effect for causal relation, condition and result for conditional relation, etc.
        \item Build a complete logic framework of the sentence. 
    \end{enumerate}
    \item \textit{Logical Structure Comparison}: Finally, GPT-4 compares the logical structure of the answer segment with the relevant context and judge if the answer segment is logically consistent with the reference. 
\end{enumerate}

The last step of FCE is the aggregation of answer-level factual consistency measurement. Specifically, an answer is marked factual consistent if and only if all of its decomposed segments have passed the above two consistency evaluation stages.

\begin{table*}
  \caption{Performance comparison of factual consistency evaluation on the synthetic dataset. }
  \label{tab:accuracy on synthetic dataaset}
  \begin{tabular}{c|c|cccccccccccccc}
    \toprule
   \multirow{2}{*}{\textbf{Method}} & \multirow{2}{*}{\textbf{Total}} & \multirow{2}{*}{\textbf{Pos.}}  & \multicolumn{9}{c}{\textbf{Negative samples}} \\ \cline{4-13}
    & & & {Hallu.} & {KCont.} & {KInve.}& {KConf.} & {KConc.}&{LOver.} & {LCaus.}& {LConf.}& {LIncl.}& {LOthe.}\\ \midrule
FACTSCORE(GPT-3.5) & 70.36 & 37.31 & 90.45 &\textbf{100} & 94.44 & 55.56 & 94.29 &78.57& 64.29 & 68.00 & 46.34 & 86.07   \\
FACTSCORE(GPT-4)   & 71.82 &33.50 & 93.97  & \textbf{100} & 96.30 & 68.25 & 97.14 & 87.5&60.71 & 72.00 & 51.22 & 88.37   \\
FELM              & 68.05 &77.67 & 42.21 & 99.27 & 91.98&22.22 & 88.57&69.64 & 42.86 & 54.00 &  4.88 & 32.56  \\
RAGAS(GPT-3.5)     & 69.59 &70.81 &  76.89 & 98.54 & 71.60 & 49.21 & 87.14&54.46 & 39.29 & 48.00 & 34.15 & 44.19 \\
RAGAS(GPT-4)       & 76.37 &73.60 & 93.97 & 99.27& 79.01 & 52.38 & 90.00 &58.93& 50.00 & 50.00 & \textbf{53.66} & 72.09 \\
RefChecker        & 78.52 & 76.14& 95.48 & \textbf{100} & 87.65 & 63.49 & 92.86 &55.36& 50.00 & 52.00 & 36.59 & 67.44 \\
L-Face4RAG (Ours)        & \textbf{93.38} &\textbf{96.19} & \textbf{96.98} & \textbf{100} & \textbf{98.77} & \textbf{76.19} & \textbf{98.57} & \textbf{90.18} &
 \textbf{92.86} & \textbf{80.00} & 51.22 &\textbf{90.70}   \\
\bottomrule
  \end{tabular}
\end{table*}

\begin{table*}[tb] \label{tab:accuracy result}
\caption{Performance comparison of factual consistency evaluation on the real-world dataset. }
\label{tab:accuracy on real-world dataaset}
\begin{tabular}{c|c|ccccccc}
\toprule
\textbf{\makecell{Method}} & \textbf{\makecell{Total}}& \textbf{\makecell{Baichuan2}} & \textbf{\makecell{ChatGLM3}} & \textbf{\makecell{GPT-3.5}} & \textbf{\makecell{GPT-4}} & \textbf{\makecell{Alpaca2 (CH)}} & \textbf{\makecell{Qwen}} \\ \midrule
FACTSCORE(GPT-3.5) & 53.33 &54.0 &55.5 & 47.5&51.5 &59.0&52.5 \\
FACTSCORE(GPT-4)   & 54.67 &55.0 &59.5 & 46.5&52.5 & 63.0&51.5\\
FELM              & 55.00 & 49.6& 56.0&56.8 &52.0 &55.6&60.0 \\
RAGAS(GPT-3.5)     & 65.92 &64.5 & 68.5&64.5 &60.0 &65.0&73.0 \\
RAGAS(GPT-4)       & 72.92 &72.5 & 74.0& 71.5& 68.5&76.5&74.5 \\
RefChecker        & 68.25 &62.0 & 72.0& 66.5& 63.0&74.5&71.5 \\
L-Face4RAG (Ours) & \textbf{87.75} &\textbf{90.0} &\textbf{88.0}&\textbf{81.5}&\textbf{86.0}&\textbf{93.5}&\textbf{87.5}\\
\bottomrule
\end{tabular}
\end{table*}

\section{Experiments}
In this section, we conduct extensive 
experiments to evaluate the effectiveness of our proposed {L-Face4RAG} pipeline. Our experiments show that on both synthetic data and real-world data in Face4RAG benchmark, our L-Face4RAG method substantially outperforms the existing FCE methods. Notably, its superiority goes beyond the Chinese RAG task from which L-Face4RAG is originally motivated, as L-Face4RAG achieves SOTA results on 6 out of 7 of the existing  English datasets and also a substantially higher average score on all tasks. 

\subsection{Experimental Setup }

\textbf{Baselines} We compare L-Face4RAG with four GPT-based fine\-/grained FCE methods:
\begin{itemize}
    \item FACTSCORE \cite{min2023factscore} first breaks the answer into a series of atomic facts and then assigns a binary label to each atomic fact individually.
    \item FELM \cite{chen2023felm} first segments the answer into fine-grained textual spans and then evaluates the factual consistency of all textual spans collectively. It outputs the corresponding numbers of factual inconsistent textual spans if existed.
    \item Ragas \cite{es2023ragas} first extracts a set of statements from the answer and then evaluates the factual consistency of all statements collectively, outputting a binary label for each statement along with the corresponding reason for the assessment.
    \item Refchecker \cite{hu-etal-2023-refchecker} extracts knowledge triplets from the answer and evaluates each knowledge triplet separately.
\end{itemize}

\textbf{Implementation Details} As the above FCE baselines are originally designed for tasks in English, we adapt them to our Chinese RAG task by translating their prompts into Chinese.

When experimenting with FELM, we utilize the Reference-doc augmented evaluator \cite{chen2023felm}, in alignment with our task which is focused on evaluating the factual consistency of answers against their references. Specifically, we input our references as the retrieved reference doc in FELM's evaluation framework. We select the best-performing estimator in \cite{chen2023felm}, i.e., decomposing the answer with segment-based method and utilizing GPT-4 as the factual error detector.

Since the original settings of FACTSCORE and RAGAS are based on GPT-3.5, we conduct experiments with both GPT-3.5 and GPT-4 to eschew the effect of the possible performance gap between GPT-3.5 and GPT-4 on the empirical results.

Finally, to apply our proposed evaluation pipeline, we decompose the answer into segments and assess the factual consistency of each segment respectively. The outputs include both the label and the corresponding explanations. To derive deterministic output from GPT-4, we set its temperature to 0.

\subsection{Performance Comparison on Face4RAG}
We first compare the performance of our proposed L-Face4RAG pipeline against various FCE baselines on the Face4RAG benchmark, which includes a synthetic dataset and a real-world dataset.

\textbf{Synthetic Dataset} 
In Table~\ref{tab:accuracy on synthetic dataaset}, we report the predictive accuracy of different error types for examined FCE methods on the synthetic dataset. From the results, we have the two main observations. (i) Our method achieves the highest accuracy on most of the error types (except on \textit{LIncl.} where it is slightly worse than RAGAS with GPT-4), which amounts to a significant improvement on overall accuracy compared to all the baselines. (ii) In particular, the performance gap between our method and baselines on error types of logical fallacy are much larger than the gap on other error types, which indicates that our method is especially capable of handling logical fallacy owing to our specific algorithmic designs. 

\textbf{Real-world Dataset} 
In Table~\ref{tab:accuracy on real-world dataaset}, we compare the predictive performance of our proposed pipeline with previous FCE methods on the real-world dataset. From the results we observe that: (i) The overall accuracy of our method is substantially higher than those of the baseline FCE methods, showing superiority in real-world scenarios. (ii) Moreover, on most of the subsets generated by different LLMs, our method consistently outperforms baseline methods, which indicates the superiority of our method is universal and independent of the error distribution, which is in line with the empirical results on the synthetic dataset.  

\subsection{Performance Comparison on Existing FCE Benchmark}

\begin{table*}
  \caption{Performance comparison of factual consistency evaluation on the existing benchmark.}
  \label{tab:additional experiment}
  \begin{tabular}{c|c|cc|cc|cc|c}
    \toprule
     \multirow{2}{*}{\textbf{Method}} & \multirow{2}{*}{\textbf{Avg.}} &  \multicolumn{2}{c}{\textbf{RAG}} & \multicolumn{2}{c}{\textbf{Summ.}} & \multicolumn{2}{c}{\textbf{Dial.}} & \textbf{Fact Verif.}\\ \cline{3-9}
    & & {RAGAS\cite{es2023ragas}} & {RefChecker\cite{hu-etal-2023-refchecker}} & {FRANK\cite{pagnoni2021understanding}} & {SummEval\cite{fabbri2021summeval}} & {$Q^2$\cite{honovich2021q}} & {DialFact\cite{gupta2021dialfact}} & {VitaminC\cite{schuster2021get}}  \\ \midrule
FACTSCORE(GPT-4)  & 70.5  & 70 &61 & 80&65&74&72&71  \\
FELM              & 74.2 & 71&63&70&\underline{82}&\underline{83}& \textbf{79}&72 \\
RAGAS(GPT-4)      & 76.9  & \underline{88} &69&\textbf{87}&80&77&69&69 \\
RefChecker        & \underline{78.4} & 86 & \textbf{73}&85&80&80&72&\underline{73}\\
L-Face4RAG (Ours) & \textbf{84.2}        & \textbf{91} &\textbf{73}&\textbf{87}&\textbf{90}&\textbf{84}&\underline{77}&\textbf{88}\\
\bottomrule
  \end{tabular}
\end{table*}

\begin{table}[tb]
\caption{Main results of the ablation studies on the synthetic dataset. We compare L-Face4RAG with the variants using conventional answer decomposition (A.D.), removing the COT (w/o COT), and removing the logic consistency evaluation (w/o logi.eval). Overall accuracy and the accuracy on positive or negative samples are reported.}
\label{tab:answer decompostion ablation result}
\begin{tabular}{ccccc}
\toprule
  & \textbf{L-Face4RAG} & \textbf{A.D.} & \textbf{w/o COT} & \textbf{w/o logi.eval} \\ \midrule
Overall & 93.38& 76.44&79.60&88.99 \\ 
-Positive & 96.19 & 91.62&51.27&97.46\\ 
-Negative & 92.15&69.83&91.93&85.30\\ 
\bottomrule
\end{tabular}
\end{table}

\begin{table}[tb]
\caption{Comparison of the accuracy between L-Face4RAG and the counterpart method with conventional answer decomposition (A.D.) or without logic consistency evaluation (w/o logi.eval) for detecting specific error types of negative samples on the synthetic dataset.}
\label{two ablation result}
\begin{tabular}{ccccc}
\toprule
  & &\textbf{L-Face4RAG}  & \textbf{A.D.} & \textbf{w/o logi. eval} \\ \midrule
Hallucination& Hallu. & 96.98 & 90.45&96.98\\
\midrule
\multirow{4}{*}{Knowledge}  &KCont. & 100.00 & 100.00& 100.00\\
&KInve.  & 98.77& 74.07&97.53\\
&KConf. & 76.19 &41.27&66.67\\
&KConc. & 98.57& 90.00&94.29\\
\midrule
\multirow{5}{*}{Logical}  &LOver. & 90.18 & 42.86&83.93\\
& LCaus. & 92.86 &32.14&35.71\\
& LConf. & 80.00 &34.00&64.00\\
& LIncl. & 51.22& 31.71&29.27\\
& LOth.  & 90.70&44.19&65.12\\ 
\bottomrule
\end{tabular}
\end{table}

We then evaluate the robustness and applicability of the proposed L-Face4RAG method on other factuality detection tasks, and in English. Specifically, we consider several commonly used FCE benchmarks in English on various tasks, including RAG\cite{es2023ragas,hu-etal-2023-refchecker}, summarization\cite{pagnoni2021understanding, fabbri2021summeval}, dialogue\cite{honovich2021q,gupta2021dialfact} and fact verification\cite{schuster2021get}.

In Table~\ref{tab:additional experiment}, we report the predictive accuracy of examined FCE methods on the above tasks. The results show that our proposed L-Face4RAG achieves SOTA results on 6 out of 7 of the existing datasets and also a substantially higher average score on all tasks, indicating the effectiveness of L-Face4RAG beyond the original factuality evaluation task in RAG, and its robustness to other languages. This validates the wide-applicability of our proposed method. 

Besides the above comparison among different methods, we also observe that the ranking of the average score of various methods on the above commonly used benchmarks is similar to the ranking of the average score on all public tasks is 0.9, and the same 0.9 between the rankings on our real-world dataset and the public datasets. This validates the strong correlation between the evaluation results of our new benchmark and the results on existing benchmarks.

\subsection{Ablation Study}

We now verify the specific design choices of our proposed evaluation pipeline by ablation study on Face4RAG benchmark. Specifically, we examine the effectiveness of each designed module by comparing L-Face4RAG with the counterpart method without such a module.
Due to space limit, here we only present the results on the synthetic dataset. Results on the real-world dataset are qualitatively similar and deferred to Appendix~\ref{app:Ablation Study Results}. 

\textbf{Evaluating the Answer Decomposition Module. (A.D.)} Recall that our decomposition module preserves the logic connection within one segment, which may help better identify logical fallacy while reducing extra hallucination induced by decomposition. To verify this point, we conduct an ablation study by replacing our approach by a conventional decomposition method \cite{es2023ragas}. As presented in Table~\ref{tab:answer decompostion ablation result} and Table~\ref{two ablation result}, we observe a severe decline of overall accuracy in the counterpart method, especially for negative samples related to logical fallacy. This phenomenon accords with our intuition that conventional answer decomposition method may fail to detect logical fallacy since some logical connections may be discarded during the decomposition. In addition, positive samples also have a slight decrease in accuracy, which justifies the third principle in our logic-preserving answer decomposition module, i.e., preserving the structure of the original answer may alleviate the introduction of extra hallucination. 

\textbf{Evaluating the Introduction of COT.(w/o COT)} Recall that COT is adopted in both stages of factual consistency evaluation, which instructs the model to conduct finer-grained fact consistency evaluation and sophisticated logic consistency evaluation, respectively. To validate the introduction of COT, we consider a counterpart method that removes the detailed steps in the instructions and requires GPT-4 to directly generate evaluation without outputting the underlying reasoning process. As presented in Table~\ref{tab:answer decompostion ablation result}, the overall accuracy drops severely after removing COT (from 93.38\% to 79.60\%), especially for the positive samples (from 96.19\% to 51.27\%). This justifies the benefit of introducing COT into FCE.

\textbf{Evaluating the Stage of Logical Consistency Evaluation. (w/o logi. eval)}
To evaluate the effect of our proposed logical consistency evaluation stage on error detection, we construct a counterpart method by removing the second stage from our pipeline. The results presented in Table~\ref{two ablation result} show that the counterpart method incurs a decline in overall accuracy. Among all the error types, logical fallacy contributes the major part of accuracy decline, which aligns with our main motivation of the second stage design for logical fallacy evaluation. In addition, there is a slight decrease in the accuracy of knowledge error. A possible reason is that the second stage may supplement the detection of some knowledge errors that are missed in the first stage. Hence, the second stage also benefits the detection of knowledge error. Note that for the hallucination error, we have not observed any obvious change in the detection accuracy; this matches our intuition that hallucination error has no relation with logical fallacy.

\section{Conclusion}
In this work, we give a systematic study of factual consistency evaluation in RAG. Specifically, we first propose a comprehensive benchmark termed Face4RAG, which includes the synthetic dataset and the real-world dataset. In light of the possible failure of existing FCE methods in detecting logical fallacy in RAG, we then propose a novel FCE method termed L-Face4RAG. Compared to previous method, our method has two novel designs, i.e., logic-preserving decomposition and fact-logic FCE, which can better characterize the logical relations in different pieces of information in the sentence, leading to higher ability of logical fallacy evaluation. Extensive experiments on both the synthetic and real-world datasets verify the effectiveness of the L-Face4RAG method. Notably, the superiority of L-Face4RAG is consistent on a wide range of factuality detection benchmarks beyond the Chinese RAG task. Elaborated ablation studies also justify our core algorithm designs. 


\clearpage

\bibliographystyle{ACM-Reference-Format}
\balance
\bibliography{sample-base}



\appendix
\section{Construction Details about Synthetic Dataset}
\label{construction of the synthetic dataset}
\textbf{Negative Samples} The negative samples are constructed based on WebCPM \cite{qin2023webcpm}, a web-enhanced question answering dataset in Chinese, following the aforementioned error typology. We use GPT-4 \cite{achiam2023gpt} to generate the negative samples and design specific prompts corresponding to each type of error. For every sample in WebCPM, we rewrite them for every error type to collect our synthetic negative sample dataset. Detailed prompts are provided at our benchmark webpage.\textsuperscript{\ref{link_face4rag}}

For the hallucination error, we construct corresponding data according to three levels of difficulty for the evaluator to detect inconsistency. To be more specific, the easiest samples, in the first group, are completely off-topic from the reference. The second group includes content that is on-topic but contains ungrounded information. The third group of sample mixes factually consistent information with hallucinated content, resulting in sentences where some parts are supported by the reference, while others are ungrounded. This mixture poses a challenge for evaluators, as it could mistakenly be labeled as "consistent" due to the presence of some consistent information.

%
For the remaining two categories, i.e., knowledge error and logical fallacy, we design a specific prompt for each error type except the Contradiction Error (\textit{KCont.}). For \textit{KCont.}, since it may occurs at different levels of granularity \cite{de2008finding}, i.e., word or sentence, we design one prompt for each level. Specifically, the prompt of the word-level \textit{KCont.} aims to select specific words in the answer and replace them with antonyms, and the prompt of the sentence-level \textit{KCont.} is designed to construct a new answer semantically contradicting the reference. Since the types of logical connections are diverse and comprehensive, for the completeness of the dataset, we consider a new error type called \textit{Other Logical Fallacy (LOthe.)}, which accounts for potential errors in some complex logical connections uncovered by our previously defined four types of logical fallacy. The prompt of \textit{LOthe.} is designed to drive GPT-4 to insert an arbitrary logical connection error into anywhere of the original answer.




\textbf{Positive Samples} To enrich the diversity of positive samples, we employ the commonly used data augmentation techniques \cite{li2022data} to generate more positive samples based on the answers from WebCPM. Our data augmentation process supplements the positive samples in WebCPM by synonym replacing and paraphrasing techniques via the prompts at the word or sentence level. Specifically, at the word level, we prompt GPT-4 to randomly replace some words in the answer with their synonyms; at the sentence level, we prompt to summarize the reference or rephrase the answer without changing the meaning of the original sentence.

\textbf{Construction Details} Following the methodology in previous research \cite{muhlgay2023generating}, we utilize the few-shot technique \cite{brown2020language}, in conjunction with the Chain of Thought (COT) \cite{wei2022chain} approach, to guide GPT-4 \cite{achiam2023gpt} to construct high-quality samples. We provide clear directions and relevant examples in the prompt and ask the model not only to produce the newly constructed samples, but also to show the thinking process behind the modifications it makes to the samples in the output. This ensures that the model is indeed generating new samples in the direction we desire. The construction prompts for both positive and negative examples are provided at our benchmark webpage.\textsuperscript{\ref{link_face4rag}}


\textbf{Human Annotation Refinement} The above construction process produces a coarse label of factual consistency for each sample. To enhance the quality of the labels, we further engage 12 human experts to annotate the factual consistency of each answer via a two-step approach \cite{min2023factscore}. Specifically, the human annotator first decompose the answer into multiple segments; for each segment, the annotator is required to judge whether it is factual consistent with the reference and give the evidence of the judgement. Then the human annotations on all segments are aggregated to yield a factual consistent label for the answer.

\section{Ablation Study Results on the Real-world Dataset}
\label{app:Ablation Study Results}
Table~\ref{tab:ablation study results on the real-world dataset} further validates the effectiveness of our approach, particularly highlighting its importance in practical scenarios where enhancing the recall of negative samples is crucial while preserving the discriminative ability of positive samples.

\begin{table}[H]
\caption{Ablation Study Results on the Real-world Dataset. Here "ours" refers to our original pipeline, "A.D." refers to the ablation result of answer decomposition, "w/o COT" refers to the ablation of COT, and "w/o logi. eval" refers to the ablation of the logical consistency evaluation.}
\label{tab:ablation study results on the real-world dataset}
\begin{tabular}{ccccc}
\toprule
  & \textbf{L-Face4FAG} & \textbf{A.D.} & \textbf{w/o COT} & \textbf{w/o logi. eval}\\ \midrule
Overall & 87.75& 76.75&65.50 &86.50\\ 
-Positive & 94.60 &82.87 &55.99 &95.65\\ 
-Negative & 75.96&66.21 &81.86& 70.75\\ 
\bottomrule
\end{tabular}
\end{table}

\section{Statistic Details about Real-World Dataset}
\label{statistic details}
Table~\ref{tab:real world dataset statistics different model} shows the statistics of 200 model-generated answers in our real-world dataset from six LLMs. Table~\ref{statistic details about Face4RAG} shows the specific information about the error distribution about the six LLMs in the real-world dataset.

\begin{table}
\caption{ Statistics of 200 model-generated answers in our
real-world dataset from six LLMs. "Avg. Length" indicates the average length of the generated answer. "Error Rate" indicates the ratio of factual inconsistent answers.}
\label{tab:real world dataset statistics different model}
\begin{tabular}{ccc}
\toprule
\textbf{MODEL} & \textbf{Avg. Length} & \textbf{Error Rate(\%)}  \\ \midrule
Baichuan2 &320.6 & 40.5 \\
ChatGLM3  & 158.0 & 36.5 \\
GPT-3.5  & 160.8 &  27.5\\
GPT-4 & 359.2 & 40.0 \\
Alpaca2 (CH) & 188.8 & 47.0  \\
Qwen  & 200.6 & 29.0\\ 
\bottomrule
\end{tabular}
\end{table}

\begin{table*}
\caption{Details about the distribution of the error types in the Real-World Dataset.}
\label{statistic details about Face4RAG}
\begin{tabular}{cccccccccccc}
\toprule
   &\textbf{Hallu.} & \textbf{KCont.} & \textbf{KInve.}& \textbf{KConf.} & \textbf{KConc.}& \textbf{LOver.}&\textbf{LCaus.}& \textbf{LConf.}& \textbf{LIncl.}& \textbf{LOthe.} & \textbf{Other Errors}\\ \midrule
Qwen&57.81&21.88&3.13&1.56&6.25&1.56&3.13&3.13&1.56&0&0\\
GPT-4&77.91&5.81&3.49&2.33&5.81&1.16&2.33&1.16&0&0&0\\
GPT-3.5&62.71&15.25&5.08&3.39&10.17&3.39&0&0&0&0&0\\
Alpaca2 (CH)&69.61&8.82&3.92&2.94&7.84&3.92&1.96&0&0.98&0&0\\
ChatGLM3&61.33&13.33&9.33&1.33&5.33&5.33&0&1.33&1.33&0&1.33\\
Baichuan2&70.59&12.94&2.35&2.35&5.33&3.53&1.18&0&1.33&0&0\\
\bottomrule
\end{tabular}
\end{table*}

\begin{table*} 
    \caption{Example of the Reading Comprehension Section in the National College Entrance Examination of China}
    \label{Example of the Reading Comprehension Section}
    \hrule
    \begin{flushleft}
    \textbf{Translated Reference:} \\
    The golden age of blue-and-white porcelain development was during the Yongle and Xuande periods of the Ming Dynasty, coinciding with Zheng He's voyages to the Western Seas, prompting us to ponder: Is it mere historical coincidence that both seafaring and porcelain craftsmanship reached their zenith at the same time? ... It was the blending of Chinese and foreign civilizations that successfully drove the transformation of Chinese porcelain from monochrome to polychrome, with blue-and-white porcelain uniquely illustrating the cultural evolution of the Ming era, serving as an example of traditional society's progression from uniformity to diversity. (Excerpted and compiled from "The Trajectory of the Rise of Ming Dynasty Blue-and-White Porcelain" by Wan Ming)\\
    ~\\
    \textbf{Task: evaluate the correctness of the following sentences:}\\
        ~\\
    \textbf{Translated Sentence 1:} Zheng He's voyages to the Western Seas stimulated the production, sales, and technological innovation of porcelain, heralding the golden age of blue-and-white porcelain development. 
        ~\\
        \textbf{Label:} \textit{correct}\\
        ~\\
    \textbf{Translated Sentence 2:} Factors such as the localization of raw materials ushered the development of blue-and-white porcelain into a new phase, at which point its evolution became unrelated to foreign cultures. \\
    \textbf{Label:} \textit{Incorrect}. \\\textbf{Error Type:} Contradiction Error\\
    ~\\
    \textbf{Translated Sentence 3:} Ming Dynasty society is often considered conservative, yet the styles of blue-and-white porcelain indicate that the society was relatively open and progressive. \\
    \textbf{Label:} \textit{Incorrect}. \\\textbf{Error Type:} Conceptual Substitution Error\\
    ~\\
    \textbf{Translated Sentence 4:} The blending of Chinese and foreign civilizations promoted the transformation of porcelain from monochrome to polychrome, thereby driving the society of the time towards a more diverse transition.\\
    \textbf{Label:} \textit{Incorrect}. \\\textbf{Error Type:} Causal Confusion Error\\
\end{flushleft}
\hrule
\end{table*}
\end{CJK*}

\end{document}